\documentclass[conference]{IEEEtran}
\IEEEoverridecommandlockouts
\usepackage{tikz}
\usetikzlibrary{shapes.geometric, arrows.meta}
\usepackage{bbm}
\usepackage{cite}
\usepackage{amsmath,amssymb,amsfonts}
\usepackage{algorithmic}
\usepackage{graphicx}
\usepackage{textcomp}
\usepackage{xcolor}
\usepackage{cleveref}
\usepackage{booktabs}
\usepackage{verbatim}
\usepackage{multirow}
\usepackage{url}
\def\BibTeX{{\rm B\kern-.05em{\sc i\kern-.025em b}\kern-.08em
    T\kern-.1667em\lower.7ex\hbox{E}\kern-.125emX}}
\begin{document}

\newcommand{\red}[1]{{\color{black}#1}}
\newcommand{\violet}[1]{{\color{black}#1}}
\newcommand{\blue}[1]{{\color{black}#1}}

\newcommand{\CR}[1]{{\color{black}#1}}

\title{Low-Latency Embedded Driver Monitoring System with a Multi-Task Neural Network}

\newcommand*{\affmark}[1][*]{\textsuperscript{#1}}
\newcommand*{\affaddr}[1]{#1} %
\newcommand*{\mailaddr}[1]{\tt{#1}}

\author{
  \IEEEauthorblockN{%
        Carmelo Scribano, Giovanni Cappelletti, Elia Giacobazzi, Giorgia Franchini, Paolo Burgio, Marko Bertogna
    }
  \IEEEauthorblockA{%
    \affaddr{University of Modena and Reggio Emilia, Italy}\\
    \mailaddr{\{name\}.\{surname\}@unimore.it}
    }
}

\maketitle

\begin{abstract}
Road traffic accidents remain a significant global concern, with the majority attributed to human factors such as driver distraction and fatigue. This study proposes a camera-based approach to derive useful indicators to assess driver attentiveness and alertness. The proposed pipeline jointly satisfies the stringent real-time requirements imposed by the critical application and minimizes the computational requirements to allow for deployment on a tight computational budget. To this end, we develop a lightweight multi-task neural network that predicts multiple indicators for the face region in a single forward pass. The developed model is integrated into a complete execution workflow to produce a real-time estimate of attentiveness, fatigue, and engagement in distracting activities.\footnote{Reference implementation at: \url{https://github.com/cscribano/MtDMS}}
\end{abstract}


\section{Introduction}
\label{sec:introduction}

Road traffic accidents continue to pose a significant threat to public safety, claiming the lives of over $1.2$ Millions each year globally. Human factors, including distraction and fatigue, are among the main causes of accidents \cite{10665-375016}, emphasizing the critical need for advanced in-vehicle monitoring systems. Although vehicle manufacturers have been implementing driver assistance systems of various types in their lineup for a few years now, Driver Monitoring System (DMS) are still not widely available, and even fewer aftermarket solution exists. DMS can be implemented with vastly different approaches, by monitoring physiological parameters of the driver or by analyzing dynamic parameters of the vehicle, such as steering or accelerations. The first category can be highly effective in estimating the alertness level, but highly invasive for the driver. The second category is unreliable due to the low correlation between the vehicle dynamics and early signs of inattentive behaviours. Camera-based solutions rely on one or multiple cameras to monitor the driver behaviour in real-time, leveraging advanced computer vision and machine learning techniques. These approaches are non-invasive and potentially highly reliable and cost-effective. Despite its potential, the main limitation of the large-scale application of camera-based DMS is the high latency due to the execution of complex computer vision pipelines consisting of multiple deep neural networks.
\begin{figure}[!h]
     \centering
    \includegraphics[width=0.45\textwidth]{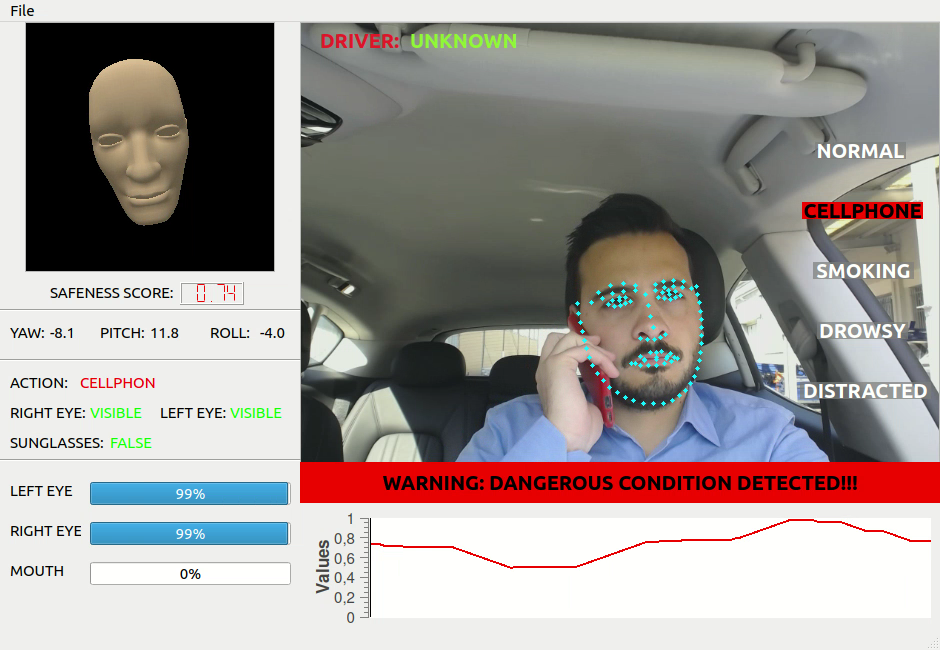}
    \caption{Interface of the developed DMS based on the proposed Multi-task network.}
    \label{fig:dms4ever}
\end{figure}
A typical DMS needs to combine several indicators to accurately determine alertness of the driver, typically: the opening of eyes, mouth, and head posture.  To this end, multi-task learning is a key paradigm to enable real-time DMS capabilities. With a carefully designed architecture and specialized training strategy, a single deep neural network can simultaneously solve multiple classification and regression tasks in a single forward pass.

In this work, we present an end-to-end driver monitoring system capable of real-time performance on a low-power edge computing unit. The core of our system is a novel Multi-Task (MT) convolutional neural network (CNN) capable of simultaneously inferring all the required indicators for the DMS algorithm, specifically:
\begin{itemize}%
    \item Regression of $98$ facial landmarks.
    \item Regression of the eyelid opening level for each eye.
    \item Classification of the visibility of each eye.
    \item Classification of the level of mouth opening.
    \item Regression of $3$ angles for head orientation (yaw, pitch, and roll).
    \item Identification of $2$ distracting actions (cell phone use or cigarette smoking).
\end{itemize}
The full pipeline includes an initial phase for extracting the region of interest tightly enclosing the driver's face and a post-processing step to temporally aggregate the indicators extracted by the MT model and produce a continuous estimate of the alertness level and fitness to drive. The full system is developed in the Nvidia Jetson ecosystem of embedded devices, evaluating both the entry-level Jetson Nano and the more powerful Xavier NX. This choice brings the clear advantage of a fully featured CUDA-capable GPU architecture, within a highly efficient System on Chip (SoC) that meets the stringent efficiency requirements of the automotive domain. We leverage the TensorRT framework for high inference performance and optionally integrate with the Robotic Operating System (ROS) for modular operation and seamless interoperability with external components.\\
\subsection{Related Works}
The research aspect involved in Driver Monitoring spans a vast set of disciplines, including Computer Vision, Edge computing, Vehicle Engineering, and Human Physiology. Existing approaches for camera-based DMS are roughly classified into two main categories. In Face-Centric approaches, the camera  is positioned in front of the driver's face,  which is the region mainly analyzed to infer signs of drowsiness or distraction. In Body-Centric models, instead, the camera observes the entire upper part of the driver's body, usually in profile, with a camera typically placed on the door pillar opposite the driver's side. The proposed system belongs to the first category.\\

\paragraph{Drowsiness Estimation}
In a seminal study on driver drowsiness \cite{wierwille1994evaluation} from 1994 the PERCLOS  metric is introduced, which to this day is still used. PERCLOS is originally defined as the percentage of the time over one minute that the eyelids are at least $80\%$ closed. This definition can be generalized as the $ \text{PERCLOS}(\%)$ which is as a function of the percentage of the time that the eye is closed \cite{junaedi2018driver}. 

\begin{equation}
    \text{PERCLOS} = \frac{\text{Total Time with Eyes Closed}}{\text{Total Monitoring Time}} \times 100\%
    \label{eq:perclos}
\end{equation}
From an implementation standpoint, a PERCLOS-based drowsiness detection system can be designed by relying on the detection of facial landmarks \cite{reddy2017real}. The eye-opening state can be derived from the measurement of the Eye Aspect Ratio (EAR), defined as the ratio between height and width of the eye. Yawning is another early indicator of drowsiness. Similar to the EAR, the Mouth Aspect Ratio (MAR) can be defined using facial landmarks detected for the mouth region. The proposed model, presented in \Cref{sec:model}, performs a direct estimation of the level of eye and mouth opening, wich allow for a more efficient estimation of EAR and MAR without relying on facial landmarks. \\

\paragraph{Multi-Task Driver Monitoring}
Only a handful of existing contributions propose multi-task learning approaches for DMS-related tasks. In \cite{celona2018multi} the authors combine classical computer vision algorithms for face analysis, implemented in the Dlib library \cite{dlib09}, with an original CNN that simultaneously classifies the state of eye-opening, mouth, head pose, and, most interestingly, the estimate of driver fatigue. In \cite{kim2019lightweight} the multi-task classification model is based on a Mobilenet architecture and is trained for the $3$ facial behavior classification tasks:  eyes (open, closed) head (front, up, down, right, left) and mouth (open, closed) and an overall label of the driver's state (normal, distraction, fatigue, drowsiness). The model  described in \cite{yang2020all} is also similar to ours. They develop a lightweight multi-task CNN (DANet) to solve the tasks of head pose estimation, landmarks detection, and gaze estimation of special interest. All of those contributions are closed source and rely on proprietary datasets, limiting the options for comparisons. Our model solves the highest number of parallel tasks ($6$), with the noticeable addition of the distraction classification tasks. Furthermore, our contribution details the complete inference pipeline, with an emphasis on inference performance and analyses of latencies on the target hardware.

\begin{figure*}[!ht]
    \centering
    \includegraphics[width=0.8\textwidth]{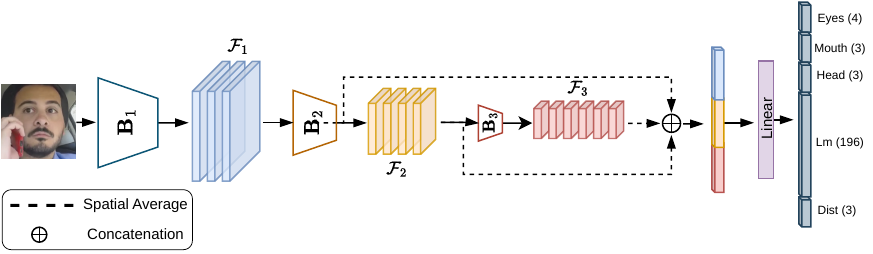}
    \caption{Architecture of the proposed Model}
    \label{fig:dmsarch}
\end{figure*}

\section{Methodology}

\subsection{Multi-task DMS Model}\label{sec:model}

The proposed Multi-Task CNN takes as input an RGB image $I\in \mathbb{R}^{(3\times w \times h)}$ depicting the region of the image that tightly 
encloses the subject's face and simultaneously predicts: 98 Facial landmarks, Eye Opening and visibility levels, Mouth Opening label, Head Orientation and Distraction Classification. The model output is a $209$ elements vector obtained as a concatenation of the individual tasks' outputs. Specifically, Face landmarks are normalized with respect to the input dimension, together with three unnormalized Euler angles (yaw, pitch, roll) for head orientation. For each eye, we regress a continuous value in $(0,1)$ representing the eyelid opening level (a level of 0 indicates a fully closed eyelid) and a binary classification label for the eye visibility (useful when eyes are blocked by sunglasses or self-occusion caused by high head rotation). A $3$ level classification is obtained with a softmax activation for the mouth opening level (closed, semi-open, open) and similarly for the driver distraction class (normal, phone use, smoking). In \Cref{tab:dms_tasks} we summarize the numerical definition and range of these indicators.
\begin{table}[!ht]
\centering
\caption{Definition of outputs. Regression (\textbf{Reg}), Binary Classification (\textbf{BC}) and Multi-class Classification (\textbf{MC}).}
    \begin{tabular}{lcccc}
    \toprule
    \textbf{Task} &  \textbf{Type} & \textbf{Shape} & \textbf{Act} & \textbf{Range} \\ \midrule
    F.Landmarks & Reg & 196 & None & $[0,1]$ \\
    Eyes Viz & BC & 2 & sigmoid & $[0,1]$\\
    Eyes Open & Reg & 2 & sigmoid & $[0,1]$\\
    Mouth & MC & 3 & softmax &  $\{0,1,2\}$\\
    Head & Reg & 3 & None & $[\text{-}\frac{\pi}{2}, \frac{\pi}{2}]$ \\
    Distr. & MC & 3 & softmax & $\{0,1,2\}$\\
    \bottomrule
    \end{tabular}
\label{tab:dms_tasks}
\end{table}

\paragraph{Model Architecture}
From an architectural standpoint, the proposed model is built on the architectural patterns of MobileNet-v2 \cite{sandler2018mobilenetv2}. This design is chosen as a trade-off between computational efficiency and performance, avoiding the use of custom or uncommon layers that could make the deployment on embedded hardware complicated.  %
MobileNet models replace Convolutional layers with depthwise separable convolution and propose the Inverted Residual pattern as the base building block. Specifically, the separable convolution replaces a standard $(n\times n)$ convolutional layer with $c$ input channels and $k$ output channels by a depthwise convolution consisting of $c$ filters of size $(n\times n\times 1)$ (one $n\times n$ filter per input channel), followed by a pointwise convolution consisting of $k$ filters of size $(1\times1\times c)$. This decomposition saves $c((k-1)*n^2 + k)$ parameters and reduces the computational complexity from $O(n^2ck)$ to $O(n^2c+ck)$. The Inverted Residual block first applies a pointwise $(1 \times 1)$ convolution to expand the input channels from $k$ to $k\times t$, where $t$ is the expansion factor. The expanded representation is then processed by a depthwise convolution followed by a pointwise projection. When the input and output dimensions match, a residual (skip) connection is added by summing the block input with its output.\\
\noindent The proposed model, schematized in \Cref{fig:dmsarch}, is defined as a stack of Inverted Residual blocks. The intermediate feature maps ($\mathcal{F}_1$, $\mathcal{F}_2$, $\mathcal{F}_3$) of two intermediate convolutional blocks ($\mathbf{B}_1$ and $\mathbf{B}_2$) and the last block ($\mathbf{B}_3$), are concatenated after reducing the spatial dimension with a global-averaging-pooling. The final output vector is obtained with a single feed-forward layer over the combined multi-scale feature vector. By varying the number of blocks, as well as the number of output channels $c$, the expansion ratio $t$, and the stride $s$ of each block, we define three architectural variants (tiny, small, large) with incremental computational cost.\\

\paragraph{Training}{

Training the multi-task CNN requires, in principle, a dataset with all tasks labeled a priori. Since no public dataset suitable for our goal is available, we resorted to augmenting the large-scale Landmark‑guided Face Parsing (LaPa) dataset \cite{liu2020new} with a small set of pseudo-labeled images for the distraction detection tasks, by adding 860 images depicting cell-phone usage and 715 representing smoking individuals. The LaPa dataset provides 22.176 images with 106 facial landmarks and 11 class pixel‑level semantic segmentation maps, which can be used, with the help of simple pre-processing steps, to derive the training labels for eye opening, eye visibility, mouth opening, and head pose estimation tasks. The remaining landmarks and the other labels are derived with a pseudo-labeling model, trained on the LaPa dataset, excluding distraction images. The deviation between the $5$ manually annotated landmarks and the corresponding pseudo-labeled ones serves as a reference for the quality of the former; this dissimilarity is then used during training to scale the loss terms for the pseudo-labeled set. The complete training setup involves advanced optimization techniques that fall beyond the scope of this work and will be discussed in future work.

}

\subsection{Other Functional Components}
\paragraph{Face Detection}

To localize faces in the input image, the system uses an SSD-based \cite{liu2016ssd} face detector designed for edge deployment\footnote{\url{https://github.com/Linzaer/Ultra-Light-Fast-Generic-Face-Detector-1MB}}. Its compact architecture enables low-latency inference, making it suitable for high-frequency behavioral analysis. The detector comes in two variants: a Slim version with a lightweight backbone composed of stacked depthwise separable convolutions, and an enhanced RFB version incorporating a Receptive Field Block \cite{Liu_2018_ECCV}, used %
to improve the trade-off between localization accuracy and computational efficiency. Before inference, input frames are linearly resized to the detector’s fixed resolution of $320 \times 240$ pixels, which provides sufficient spatial detail while maintaining high throughput. To minimize end-to-end latency, the Non-Maximum Suppression (NMS) step used to remove highly overlapping detections is embedded directly in the model graph and executed using a high-performance TensorRT implementation. The impact of this choice is analyzed in \Cref{tab:facedet_deploy}.\\ %

\paragraph{DMS Heuristics}

The system employs specific heuristics to assess the driver's state, tuned to provide an accurate evaluation of safety conditions. Each potential source of impairment is associated with a variable serving as an indicator of its magnitude. These indicators are assigned to an integer $S \in \{0, 1, 2\}$, representing increasing danger levels. Monitoring is performed by comparing the variables against two-tier statistical thresholds, 
$\tau_{low}$ and $\tau_{high}$:
$S_{\text{module}} = [\text{metric} > \tau_{low}] + [\text{metric} > \tau_{high})]$ where $[\cdot]$ is the Iverson bracket.
Each $S_{\text{module}}$ contributes to the overall safeness score of the system:
\begin{IEEEeqnarray}{rCl}
\text{Safeness Score} & = & \lambda_1 S_{\text{perclos}} - \lambda_2 S_{\text{mouth}} - \nonumber \\
& & \lambda_3 (1 - S_{\text{head}}) - \lambda_4 (1 - S_{\text{action}})
\label{eq:safeness-score}
\end{IEEEeqnarray}
where $S_\text{perclos}$, $S_\text{mouth}$, $S_\text{head}$, and $S_\text{action}$ represent the risk scores of the respective behavioral indicators, and $\lambda_\text{i}$ denotes the weight assigned to each contribution. The individual risk scores are derived from the raw metrics as follows: $S_\text{perclos}$ is computed according to \eqref{eq:perclos}, while $S_\text{mouth}$ and $S_\text{action}$ are obtained by comparing the respective detection frequencies against predefined thresholds. The head orientation score, $S_\text{head}$, is defined as the normalized deviation from a calibrated zero-point reference. Finally, these indicators are aggregated into the global safeness score using the weighted contribution model defined in \eqref{eq:safeness-score}.

\subsection{Finite State Machine}
\begin{figure*}[!ht]
    \centering
    \includegraphics[width=0.9\textwidth]{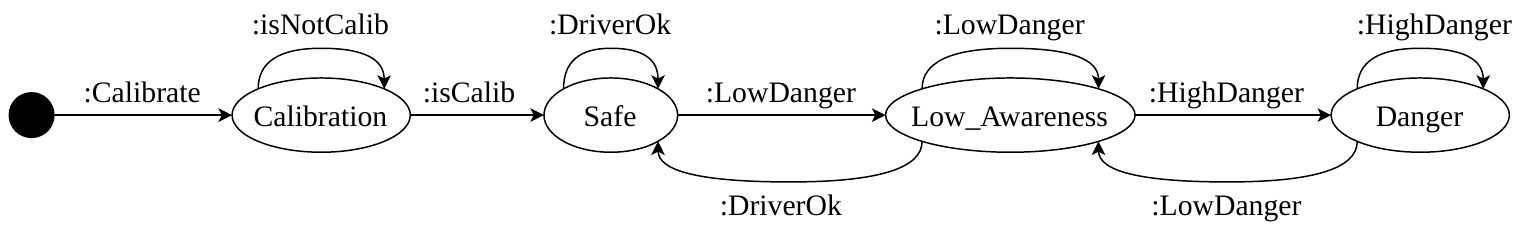} 
    \caption{Finite State Machine representing the DMS state transition logic.}
    \label{fig:dms_fsm}
\end{figure*}

The functional pipeline of the DMS can be conceptualized as a hierarchy of Finite State Machines, specifically comprising a perception FSM and a dedicated DMS FSM; the former is tasked with processing raw images acquired by a camera to extract the behavioral features which subsequently drive the state transitions and interface updates of the latter.
\paragraph{Perception Module}
Initially, the input frame is processed by a face detection network to define a Region of Interest encompassing the subject's face, %
which will be used by an algorithm based on SORT \cite{Bewley_2016} %
to track the facial region throughout upcoming frames. %
This approach renders the need for per-frame face detection unnecessary, significantly mitigating the computational overhead to enhance real-time efficiency.     
Subsequently, the multi-task architecture ingests the cropped face ROI to output the spatial coordinates of 98 facial landmarks, alongside head orientation vectors, %
state recognition for the eyes (detecting visibility and closure) and the mouth aperture %
, as well as action classification. %
\paragraph{Decision Unit}
Upon aggregating the data from the multi-task model, the subject's conditions are evaluated against specific parameters finely-tuned to serve as reliable indicators of drowsiness and distraction. First, a preliminary eye-openness calibration is performed to establish the baseline EAR for the specific driver, along with the acquisition of a baseline head pose to serve as a zero-reference coordinate system; %
These steps are crucial to account for inter-individual variability in ocular morphology and to quantify subsequent head angular displacements%
\footnote{The calibration phase operates under the assumption that the initial baseline is established while the subject is in an alert and non-impaired state.}. Following the setup, the application initiates real-time monitoring by computing the PERCLOS index within a fixed time window, %
 assessing in parallel the mouth state to detect yawnings, thereby providing a multi-modal estimation of the subject's drowsiness. 
In addition to fatigue monitoring, the DMS simultaneously evaluates cell phone usage and smoking activities. Lastly, the driver's head rotation factor is rigorously analyzed to detect potential distraction events, such as prolonged gaze diversion from the roadway—a critical factor widely recognized in contemporary literature as a leading cause of traffic accidents.  

The logic governing the DMS state transitions is formally represented by the Finite State Machine illustrated in \Cref{fig:dms_fsm}. The system cycles through four primary states based on the evaluation of real-time behavioral metrics:
\begin{itemize}
    \item Calibration - The initial state in which the baseline physiological parameters are established. %
    \item Safe - The nominal operating state. This status is maintained as long as the driver's metrics remain within safety thresholds.
    \item Low Awareness - Triggered by moderate deviations from the baseline, representing early signs of fatigue or distraction.%
    \item Danger - The highest alert level, reached when critical impairment or prolonged distraction is detected. %
\end{itemize}
The system architecture follows a "worst-case" arbitration logic: %
the global state $S_{\text{global}}$ is determined by the maximum alert level identified across all monitoring modules, %
ensuring that any single critical impairment is immediately flagged. %
The root cause is recognized by isolating the module that triggered the maximum alert level; this diagnostic capability allows for contextualized intervention and more informative feedback to the user. 
$S_{\text{global}}$ %
is derived as the supremum of all individual state levels:
\begin{IEEEeqnarray}{rCl}
S_{\text{global}} & = & \max ( S_{\text{perclos}}, S_{\text{mouth}}, S_{\text{headpose}}, \nonumber \\
           &   & S_{\text{cellphone}}, S_{\text{smoking}} )
\end{IEEEeqnarray}
Furthermore, a temporal hysteresis mechanism is implemented to prevent rapid flickering between alert levels.%
Transitions in state towards a lower risk level are allowed only after a predefined cooling period $\Delta t_{cool}$, to ensure that the driver has consistently regained a safe posture before the alert is attenuated.

\section{Deployment and Evaluation}
\begin{table}[!ht]
    \centering
    \caption{Comparison between NVIDIA Jetson Nano and Jetson Xavier NX}
    \begin{tabular}{@{}lcc@{}}
    \toprule
    \textbf{Feature} & \textbf{Jetson Nano} & \textbf{Jetson Xavier NX} \\ \midrule
    GPU               & Maxwell & Volta \\
    CUDA Cores        & 128                  & 384 \\
    CPU               & 4-core ARM Cortex-A57 & 8-core ARM Cortex-A57 \\
    CPU Clock Speed   & 1.43 GHz & 1.9 GHz \\
    RAM               & 4 GB LPDDR4          & 8 GB LPDDR4 \\
    Tensor Cores      & No       & Yes (48) \\
    AI Performance    & 0.47 TOPS  & 21 TOPS \\
    Power Consumption & 5-10W                & 10-15W \\
    \bottomrule
    \end{tabular}
    \label{tab:nanovsnx}
\end{table}

\subsection{Implementation Details}
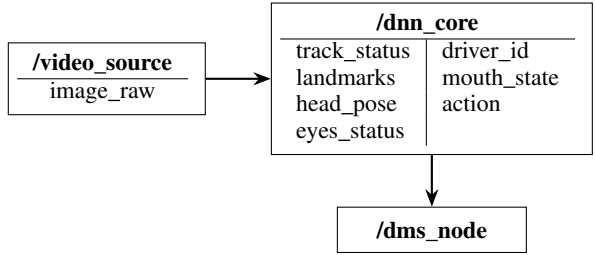
\begin{figure}[!ht]
    \centering
\begin{tikzpicture}[node distance=1.5cm, every node/.style={fill=white, font=\small}, align=center,
topic/.style={rectangle, draw, fill=white, minimum width=2.5cm, minimum height=0.6cm},
myarrow/.style={-Stealth, thick}
]
  \node (raw) [topic, xshift=-5cm] {
  \begin{tabular}{c}
    \textbf{/video\_source} \\ \hline
    image\_raw
  \end{tabular}
  };
  
  \node (t1) [topic, right of=raw, xshift=2.8cm] {
\begin{tabular}{l | l}
\multicolumn{2}{c}{\textbf{/dnn\_core}} \\ \hline

track\_status & driver\_id \\
landmarks     & mouth\_state \\
head\_pose     & action \\
eyes\_status
\end{tabular}
  };
  \node (dms) [topic, below of=t1, yshift=-0.5cm] {\textbf{/dms\_node}};

  \draw [myarrow] (raw) -- (t1);
  \draw [myarrow] (t1) -- (dms);
\end{tikzpicture}
\caption{Computational graph of the ROS-based architecture. Bold labels indicate node names; relative published topics are listed beneath the horizontal separators.}
\label{fig:ros-nodes}
\end{figure}

Following the architectural choices outlined in ~\Cref{sec:introduction}, two NVIDIA Jetson boards were evaluated as the target hardware for the DMS deployment (see \Cref{tab:nanovsnx}). These platforms were selected for their optimal balance of computational capabilities and SWaP-C constraints—Size, Weight, Power, and Cost—making them particularly suitable for developing and testing applications in resource-constrained automotive environments. The communication backbone is managed via the Robot Operating System (ROS), enabling the decoupled modular architecture depicted in \Cref{fig:ros-nodes}. This approach ensures an efficient streaming of high-frequency data between components by exploiting a publisher-subscriber paradigm. Specifically, a camera node broadcasts raw frames to a dedicated topic, which is subsequently consumed by the inference engine node. The resulting model outputs are then aggregated and published for the decision unit to evaluate. This asynchronous design ensures that each module operates at its optimal frequency, preventing computational bottlenecks across the pipeline.

\subsection{Task Performance}
First, we assess the performance of the Multi-task model introduced in \Cref{sec:model}. We evaluate the common literature metrics for each sub-task. For the landmarks regression, the Normalized Mean Error (NME), defined in (\ref{eq:nme}), computes the average Euclidean distance between the predicted facial landmarks and the ground truth annotations, normalized with respect to the interocular distance (IOD).
\begin{equation}
    \text{NME}(p,g) = \frac{1}{N} \sum_{i=1}^{N} \frac{\| \mathbf{p}_i - \mathbf{g}_i \|}{\text{IOD}(\mathbf{g})}
    \label{eq:nme}
\end{equation}
For the eye-related binary classification tasks (openness and visibility), accuracy is computed per output, and a support-weighted average over the four predictions is reported as a single score. For the multi-class mouth classification task, class-wise accuracies are computed and combined using a weighted average. For head orientation estimation, the circular error (in degrees) is reported separately for yaw, pitch, and roll. %

\begin{table}[!ht]
\centering
\caption{Comparison of model configurations}
    \begin{tabular}{lcccc}
    \toprule
    \textbf{Model} &  \textbf{NME ($\downarrow$)} & \textbf{Eyes ($\uparrow$)} & \textbf{Mouth ($\uparrow$)} & \textbf{Head ($\downarrow$)} \\\midrule
    Tiny & 3.815 & 0.955 & 0.899 & (3.855, 2.975, 3.604) \\
    Small & 2.350 & 0.978 & 0.935 & (2.622, 2.793, 3.186) \\
    Large  & 2.163 & 0.983 & 0.983 & (2.078, 2.323, 2.458) \\
    \bottomrule
    \end{tabular}
\label{tab:model_conf}
\end{table}
For the distraction classification tasks, the limited number of samples available does not allow for the isolation of a test set. Instead, we performed qualitative evaluations in a real-world deployment scenario.  Saliency maps generated with the  Score‑CAM method \cite{wang2020score} were also evaluated to analyze the model response. The details of these analyses will be discussed in future work.

\subsection{Inference Performance}

In this section, we discuss the observed latencies for the proposed Multi-task CNN model. Both boards run the same models under identical software conditions, ensuring a fair comparison of inference performance. The latency is measured on the target platform using the built-in \texttt{trtexec} tool; the inference time is averaged across $1,000$ inference passes with a $200$ ms warmup time. CPU and GPU clock frequencies are set to the maximum value to ensure reproducibility of the experiments, avoiding possible deviations induced by the platforms' dynamic power management. In addition to wall‑clock latency, the total number of model parameters is assessed, and Multiply‑Accumulate operations (MACs) are reported in Giga‑operations per second (GMACS) as a hardware‑agnostic proxy of computational cost. The results are reported in \Cref{tab:dms_deploy}.

\begin{table}[!t]
\caption{Deployment Evaluation of Multi-Task CNN in three different configurations (small, base, and large).}
\centering
    \begin{tabular}{lcccccc}
    \toprule
\multirow{2}{*}{\textbf{Model}} & \multirow{2}{*}{\textbf{Par}} & \multirow{2}{*}{\textbf{GMACs}} & \multicolumn{2}{c}{\textbf{Nano (ms)}} & \multicolumn{2}{c}{\textbf{NX (ms)}} \\
    \cmidrule{4-7} 
    & & & FP32 & FP16 & FP32 & FP16 \\ \toprule
    Tiny  & 124,566  & 76.61 & 4.91 & 4.92 & 1.25 & 0.83 \\
    Small  & 705,422  & 447.23  & 13.90 & 13.64 & 3.52 & 1.73 \\
    Large & 2,330,318 & 2400 & 75.45 & 75.67  & 15.64  & 6.71  \\
    \bottomrule
    \end{tabular}
\label{tab:dms_deploy}
\end{table}

\subsection{Latency Breakdown}
\begin{table}[!t]
\caption{Deployment Evaluation of Face Detection Model in two variants (SLIM and RFB). The NMS column indicates whether the Python/Numpy (PY) implementation or the TensorRT Plugin (RT) of Non-Maxima suppression is used.}
\centering
    \begin{tabular}{lccccccc}
    \toprule
    \multirow{2}{*}{\textbf{Model}} & \multirow{2}{*}{\textbf{Par}} & \multirow{2}{*}{\textbf{GMACs}} & \multirow{2}{*}{\textbf{NMS}} & \multicolumn{2}{c}{\textbf{Nano (ms)}} & \multicolumn{2}{c}{\textbf{NX (ms)}} \\
    \cmidrule{5-8}
    & & & & FP32 & FP16 & FP32 & FP16 \\ \midrule
    \multirow{2}{*}{SLIM}  & \multirow{2}{*}{258k}  & \multirow{2}{*}{84.43} & PY & 46.29 & 46.31 & 33.93  & 33.41 \\
                        &   &  & RT & 5.40 & 5.43 & 1.51  & 1.01 \\ \cmidrule{1-8} 
    \multirow{2}{*}{RFB}  & \multirow{2}{*}{274k}  & \multirow{2}{*}{103.57} & PY & 47.02 & 46.73 & 34.15  & 33.61 \\
                        &   &  & RT & 6.114 & 5.90 & 1.729  & 1.12\\
    \bottomrule
    \end{tabular}
\label{tab:facedet_deploy}
\end{table}

The system's performance was benchmarked by analyzing the latency contributions of each architectural component. Specifically, the evaluation focused on the individual inference times of the face detector and the DMS MTL network, as well as the overall end-to-end latency. Models were tested using both $32$-bit and $16$-bit floating-point precision. 

\paragraph{MTL network}
The multi task network inference times were evaluated over different model sizes, a detailed presentation of the results is given in \Cref{tab:dms_deploy}. The network was benchmarked with $160\times160$ images, using a batch size of 1.

\paragraph{Face Detector}
The face detector's performance was evaluated on both Jetson platforms. The analysis accounted for different Non-Maximum Suppression (NMS) implementations, specifically contrasting a standard Numpy-based approach with an optimized TensorRT layer injected in the model's graph. The input of the face detector is $320\times240$, batch size is 1.  Refer to \Cref{tab:facedet_deploy} for an accurate analysis.

\paragraph{End-to-End Latency}
  
In addition to individual model inference times, the overall end-to-end latency of the DMS was measured to assess the system’s real-time execution capabilities. Specifically, four configurations of the driver tracker module were evaluated, varying the face detector's inference frequency. In this setup, N=1 denotes execution on every frame, while N=2,4,8 signify that the network is triggered every second, fourth, or eighth frame, respectively. The Small version of the MTL model was selected as the baseline due to its optimal trade-off between accuracy and inference speed. Regarding the face detector, the RFB variant was employed throughout the benchmark to ensure consistent performance evaluation. The end-to-end latency breakdown for this configuration, under varying detection intervals N, is reported in \Cref{table:latency_breakdown}. It should be noted that the acquisition delay of the camera (50 ms at 20 FPS) was excluded from this analysis, as it represents a fixed hardware constraint that does not reflect the computational performance of the software pipeline.

\begin{table}[!t]
\renewcommand{\arraystretch}{1.3}
\caption{End-to-End DMS Latency vs. Face Detection Interval. Latency assessed on Nvidia Xavier NX.}
\label{table:latency_breakdown}
\centering
\begin{tabular}{lcccc}
\toprule
\textbf{Detection Interval} & \multicolumn{4}{c}{\textbf{Latency (ms)}} \\
\cmidrule(r){2-5}
Every $N$ frames & $N=1$ & $N=2$ & $N=4$ & $N=8$ \\
\midrule
Total Latency & 23.73 & 19.94 & 16.76 & 16.46 \\
\bottomrule
\end{tabular}
\end{table}

\section{Conclusions and Future Works}
We presented a complete pipeline for a camera-based Driver Monitoring system designed for on-edge execution on low-end embedded devices. The proposed Multi-task CNN enables the simultaneous inference of multiple biometric indicators associated with driver drowsiness and distraction, achieving strong performance across the individual tasks (\Cref{tab:dms_tasks}) while reducing overall inference latency (\Cref{tab:dms_deploy}). The full pipeline, formalized as a finite-state machine (\Cref{fig:dms_fsm}), was implemented with optional ROS support. Experimental results show that the end-to-end execution time (\Cref{table:latency_breakdown}) meets the latency requirement imposed by the acquisition rate. Future work will investigate in greater detail the training strategies adopted to address the scarcity of annotated data for distraction-related tasks. Additionally, we plan to evaluate the system on embedded platforms beyond Nvidia hardware. The increasing availability of cost-effective heterogeneous platforms equipped with dedicated inference accelerators (NPUs) represents a promising direction to further improve efficiency and maintain real-time guarantees, particularly under resource contention scenarios.

\section*{Acknowledgments}
The work is supported by the Chips joint Undertaking and its members, including the top-up funding by the  national Authorities of Germany, Belgium, Spain, Finland, Netherlands, Austria, Italy, Greece, Latvia, Lithuania  and Turkey, under grant agreement number 101139996-2. Co-funded by the European Union.

\bibliographystyle{IEEEtran}
\bibliography{main}

\vspace{12pt}

\end{document}